\documentclass[runningheads]{llncs}
\usepackage[T1]{fontenc}
\usepackage{graphicx}
\usepackage{amsmath}
\usepackage{booktabs}
\usepackage[verbose]{placeins} 
\usepackage{mathtools}
\usepackage{xcolor}
\usepackage{ctable} % Add this to your preamble
\usepackage{enumitem}
\usepackage{hyperref}
\newcommand{\filledredcirc}{\ensuremath{{\color{blue}\bullet}\mathllap{\circ}}}

\begin{document}
\setlength{\textfloatsep}{6pt}
\setlength{\intextsep}{6pt}
\setlist[itemize]{topsep=2pt,itemsep=0pt,leftmargin=*}

\bibliographystyle{splncs04}
\title{Learning to Get Up Across Morphologies: Zero-Shot Recovery with a Unified Humanoid Policy}
\author{Jonathan Spraggett\inst{1}\orcidID{0009-0007-3366-2894} }
\titlerunning{Learning to Get Up Across Morphologies}% Part of RIGHT running header

\authorrunning{J. Spraggett}
\institute{University of Toronto, 55 St. George St., Toronto, M5S 0C9, Ontario,
Canada
\email{jonathanspraggett@gmail.com}}
\maketitle              % typeset the header of the contribution
\begin{abstract}
Fall recovery is a critical skill for humanoid robots in dynamic environments such as RoboCup, where prolonged downtime often decides the match. Recent techniques using deep reinforcement learning (DRL) have produced robust get-up behaviors, yet existing methods require training of separate policies for each robot morphology. This paper presents a single DRL policy capable of recovering from falls across seven humanoid robots with diverse heights (0.48–0.81 m), weights (2.8–7.9 kg), and dynamics. Trained with CrossQ, the unified policy transfers zero-shot up to 86 ± 7\% (95\% CI [81, 89]) on unseen morphologies, eliminating the need for robot-specific training. Comprehensive leave-one-out experiments, morph scaling analysis, and diversity ablations show that targeted morphological coverage improves zero-shot generalization. In some cases, the shared policy even surpasses the specialist baselines. These findings illustrate the practicality of morphology-agnostic control for fall recovery, laying the foundation for generalist humanoid control. The software is open-source and available at: \href{https://github.com/utra-robosoccer/unified-humanoid-getup}{https://github.com/utra-robosoccer/unified-humanoid-getup}.

\end{abstract}
\section{Introduction}
Humanoid robots frequently encounter falls during operation, especially in dynamic environments such as RoboCup soccer, where collisions, balance loss, or unexpected terrain can force a robot to fall \cite{Saeedvand_Jafari_Aghdasi_Baltes_2019}. In these moments, the ability to recover quickly and autonomously is mission-critical. A robot that cannot get up is effectively out of play \cite{robocup2025rules}. Fall recovery is, therefore, not just a safety feature but a core skill required for sustained autonomy and competitive success.

However, recovering from a fall is a challenging task that differs significantly from locomotion. It involves dynamic whole-body interactions with the ground and requires precise coordination between limbs \cite{mdpi2023robocup}. Traditionally, fall recovery policies are handcrafted per robot using key frame-based (KFB) sequences, which are labor-intensive to tune and fragile under unseen initial conditions or robot variations \cite{mdpi2023robocup}. Recent work has begun to explore deep reinforcement learning (DRL) for more adaptable fall recovery\cite{rl}, but these policies are still trained and deployed on a single morphology, limiting their reuse across platforms.

In contrast to the increasing availability of multi-robot locomotion controllers such as URMA~\cite{bohlinger2025policyrunallendtoend}, no prior work has demonstrated a single get-up policy that transfers across multiple humanoid robots. Controllers like HoST~\cite{huang2025learninghumanoidstandingupcontrol} achieve robust recovery for a specific robot, but do not address generalization to other embodiments. Morphological differences, such as joint configurations, limb lengths, and torque limits, present a major challenge to sharing a recovery strategy across robots.

This paper asks: Can a single DRL policy trained across multiple humanoid morphologies learn to perform fall recovery and zero-shot transfer to unseen robots without retraining? Does increasing morphological diversity during training encourage generalizable strategies, allowing the learned controller to adapt to novel robot geometries?

To test this hypothesis, we constructed a shared observation and action space and trained a unified policy across seven different humanoid morphologies in MuJoCo.

\textbf{The key contributions are:}
\begin{itemize}
    \item Presentation of the first unified DRL policy for zero-shot fall recovery in seven humanoid morphologies\footnote{Open-source: \url{https://github.com/utra-robosoccer/unified-humanoid-getup}}.
    \item Demonstration that increasing morphological diversity during training improves generalization to unseen robots.
    \item Leave-one-out and morphological scaling experiments to analyze zero-shot generalization trends.
    \item Highlights failure cases of single-morph policies and demonstrates how shared policies overcome them.
\end{itemize}

Our results show that shared policies match or outperform per-robot baselines on zero-shot tests. This work demonstrates a path toward general-purpose, morphology-agnostic control. Reducing the cost of new robot deployment and laying the groundwork for more generalist humanoid skills in the future.

\section{Background \& Related Work}
Classical approaches often rely on predefined sequences such as KFB methods or Model Predictive Control (MPC), which require substantial tuning and may struggle with unexpected scenarios \cite{mdpi2023robocup}. For example, the RoboCup Kid-Size 2023 champion, Rhoban, employed meticulously designed KFB for the Sigmaban humanoid, which provided reliability but limited generalization \cite{gaspard2025frasa}.

Recent research has turned toward DRL as a more adaptive solution. Yang et al. \cite{10138662} utilized DRL and contact transition graphs to learn robust get-up behaviors across humanoids and quadrupeds. Building on this, Fall Recovery and Stand Up agent (FRASA) \cite{gaspard2025frasa} integrated fall detection and recovery into a unified DRL policy, significantly improving robustness over scripted approaches. However, these DRL methods remain specific to individual robots, requiring separate training for each new morphology.

Expanding this research area, generalizing robot behaviors across multiple morphologies has recently gained traction. Early methods like NerveNet \cite{wang2018nervenet} used Graph Neural Networks (GNNs) to represent robot morphologies, enabling zero-shot transfers between morphologically similar platforms. The Unified Robot Morphology Architecture (URMA) \cite{bohlinger2025policyrunallendtoend} and ModuMorph \cite{xiong2023universalmorphologycontrolcontextual} generalized locomotion control by using morphology-agnostic encoding and hyper network-based contextual modulation. Yet, these approaches typically incorporate explicit structural information, such as joint graphs or morphology vectors, into the policy. In contrast, our method is entirely morphology-blind and addresses the humanoid fall recovery. Our unified DRL policy successfully generalizes zero-shot to multiple unseen humanoid morphologies, significantly surpassing prior works that either focused solely on locomotion or relied on morphology-specific training.

Sim-to-real transfer is another crucial consideration in deploying DRL policies. Techniques such as extensive domain randomization, introduced by Tan et al. [15], are widely used. \cite{tan2018simtoreallearningagilelocomotion}. HoST \cite{huang2025learninghumanoidstandingupcontrol} also achieved robust sim-to-real transfer for a single humanoid using curriculum training and motion smoothing. FRASA demonstrated effective sim-to-real transfer for humanoid fall recovery in RoboCup competitions \cite{gaspard2025frasa}. Our approach similarly incorporates extensive domain randomization to ensure robust transferability.

Our priority is morphology-agnostic fall recovery, but realism could be boosted with Adversarial Motion Priors (AMP) \cite{Peng_2021}. The author previously used AMP with curriculum learning to produce human-like kicking, walking, and jumping on one Kid-Size robot \cite{Spraggett23}. Those policies were morphology-specific and did not cover fall recovery. Applying AMP to refine recovery motions remains an attractive future work.

\section{Methods}
To handle multiple morphologies, the single-robot approach FRASA~\cite{gaspard2025frasa} was modified: with an expanded observation space, a morphology-agnostic reward function, and a setup that trains across various robot models. This ensures that the learned policy can generalize beyond any one morphology.

\subsection{Humanoid Robot Suite}
\begin{figure}[t]
  \centering
  \includegraphics[width=12cm]{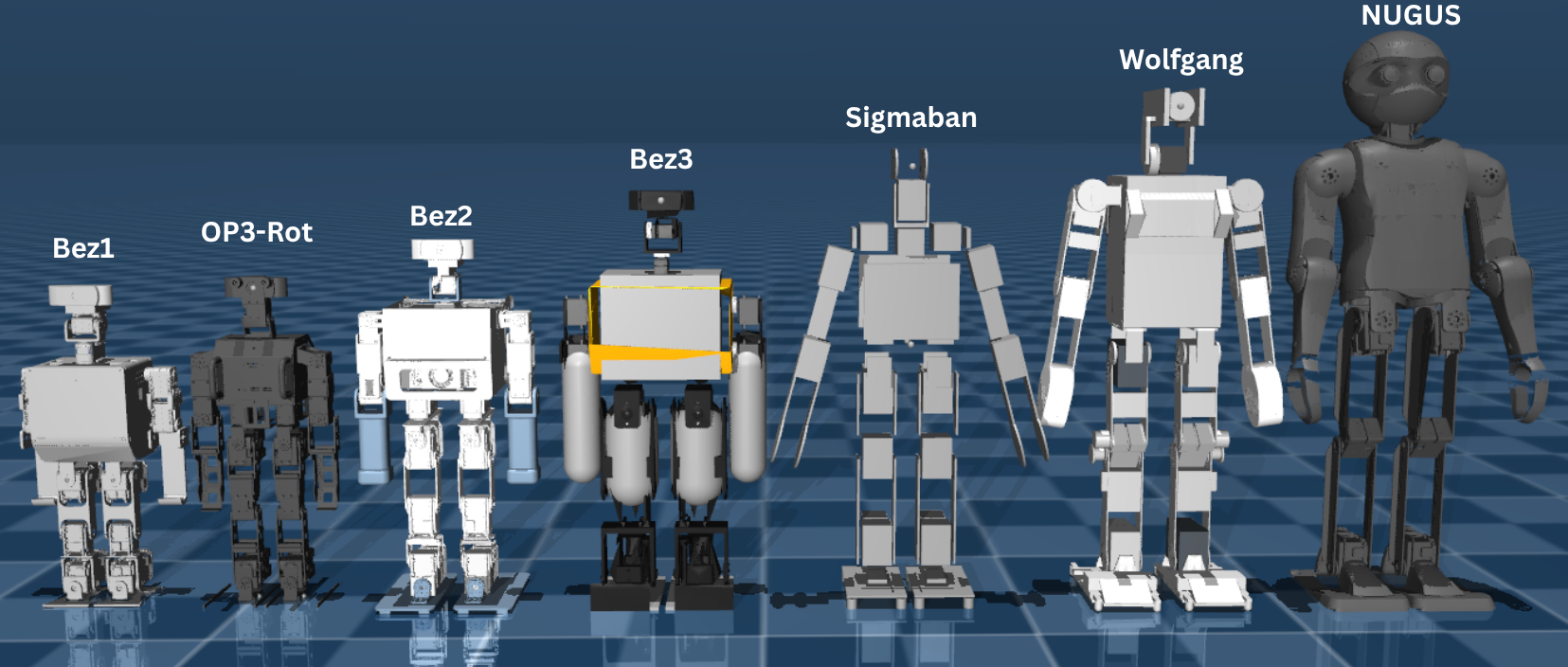}
  \caption{Visual of diverse humanoid morphologies. Ordered by size (left: smallest, right: largest). Despite differences in height (0.48m to 0.81m) and weight (2.8kg to 7.9kg), all share a similar bipedal structure, allowing a common control policy to be applied.}
  \label{fig:all_robot}
\end{figure}

Seven humanoid robot models~\cite{robocup2025opensource} were selected, as seen in Figure~\ref{fig:all_robot} and described in Table~\ref{tab:robots}. The models were converted to MJCF, MuJoCo’s native XML format, to ensure compatibility with the MuJoCo simulator. To ensure consistency across models, (i) the elbow joints on OP3 were rotated to move along the pitch axis. (ii) Adjusted initial joint angles for the hip and ankle pitch motors on Wolfgang and NUGUS to prevent immediate instability. (iii) Standardized naming conventions for body links and actuators. Additional inertial measurement unit (IMU) and foot-frame reference points were introduced to provide consistent sensor locations across all morphologies. These adjustments ensured that all robots could be controlled under a common action space and compared fairly, paving the way for training a unified policy.

% \vspace{-4pt}

\begin{table}[ht]
\centering
\begin{minipage}[t]{0.48\linewidth}\centering
\caption{Physical Description of Humanoid Morphologies}
\label{tab:robots}
\begin{tabular}{lccc}
\toprule
Robot ID & Height (m) & DoF & Mass (kg) \\
\midrule
Bez1     & 0.48 & 18 & 2.82 \\
OP3-Rot  & 0.49 & 20 & 3.15 \\
Bez2     & 0.54 & 20 & 3.86 \\
Bez3     & 0.62 & 18 & 3.18 \\
Sigmaban (Sig) & 0.67 & 20 & 7.80 \\
Wolfgang & 0.77 & 20 & 6.12 \\
NUGUS    & 0.81 & 20 & 6.68 \\
\bottomrule
\end{tabular}
\end{minipage}\hfill%
\begin{minipage}[t]{0.48\linewidth}\centering
% \begin{table}[ht]
\centering
\caption{Observation State Vector}
\begin{tabular}{lll}
\hline
\textbf{State} & \textbf{Description} & \textbf{Dim} \\
\hline
$q_t$ & Joint Positions & 5x1 \\
$\dot q_t$ & Joint Velocity & 5x1 \\
$q^{\text{des}}_{t}$ & Desired Joint Positions & 5x1 \\
$ \theta^{\mathrm{rpy}}_t$ & Trunk Euler Angles & 3x1 \\
$\dot\theta^{\mathrm{rpy}}_t$ & Trunk Angular Velocity & 3x1 \\
$h^{\text{head}}_t$ & Head Height & 1x1 \\
$a_{t-1}$ & Previous Action & 5x1 \\
\hline
\end{tabular}
\label{tab:obs}
% \end{table}
\end{minipage}
\end{table}

\subsection{Unified Action Space}
A morphology-independent action space was defined, following the scheme of FRASA \cite{gaspard2025frasa}. The policy outputs desired joint angle change, as seen in Equation \eqref{eqn:0}, for the shoulder, elbow, hip, knee, and ankle pitch joints that are common across all robots.
\begin{equation}
 \label{eqn:0}
a_{t} = \dot{q}^{desired}_t
\end{equation}

In each 50 ms control step $t$, the next target joint angles are calculated by $q^{desired}_{t+1} = q^{desired}_t + \dot{q}^{desired}_t \Delta t$. This reduced joint set and symmetric control simplify the learning problem and ensure that the action space is consistent across morphologies. The simulator enforces joint limits, so the policy does not need robot-specific scaling.

\subsection{Extended Observation Space}
The morphology-agnostic observation space expands on FRASA~\cite{gaspard2025frasa} to capture the robot’s state in a generalized way, as can be seen in Table \ref{tab:obs}. The trunk's Euler angles $\theta^{\mathrm{rpy}}_t$ and rate of change $\dot\theta^{\mathrm{rpy}}_t$ were extended to cover the roll, pitch and yaw axes. This allows the policy to determine orientation in any direction. To incentivize the policy towards standing, the vertical height of the robot's head $h^{head}$ above the feet was added as a morphology-independent metric. If the head drops below the foot, $h^{head}$ is set to a low value to penalize inverted postures. Notably, we do not include any explicit morphological identifiers. The policy must infer the necessary differences from the state dynamics alone, an approach that contrasts with methods that provide a morphology descriptor~\cite{bohlinger2025policyrunallendtoend}\cite{xiong2023universalmorphologycontrolcontextual}. This design tests the policy's ability to generalize across all embodiments.

\subsection{Morphology-Agnostic Reward Structure}
The reward function uses common physical criteria that apply to any humanoid, rather than the desired joint angles to a morphology-specific pose like FRASA~\cite{gaspard2025frasa}. The reward is described by Equation \eqref{eqn:2} and Table \ref{tab:reward}.
\begin{align}
 \label{eqn:2}
R = R_{Up} + R_{Pitch} + R_{vel} + R_{var}+ R_{collision}
\end{align}
\begin{table}[ht]
\centering
\caption{Reward Function Components and Formulas}
\begin{tabular}{ll}
\hline
\textbf{Reward Component} & \textbf{Equation} \\
\hline
Upright posture reward & $R_{Up}=\exp\left(-10\cdot\|h^{\text{head}}_t - 1.0\|^2\right)$ \\
Pitch alignment reward &$R_{Pitch}=\mathbf{1}\cdot\left[h^{\text{head}}_t > 0.4\right]\cdot\exp\left(-10\cdot\|\theta^{\mathrm{pitch}}_t\|^2\right)$ \\
Velocity Reward & $R_{vel}=0.1 \cdot \exp(-\|\dot q_t\|)$\\
Action Variation Reward & $R_{var}=0.05 \cdot \exp(-\|a_t - a_{t-1}\|)$\\
Self Collision Reward & $R_{collision}= 0.1 \cdot \exp(\text{-selfCollision}_t)$\\
\hline
\end{tabular}
\label{tab:reward}
\end{table}
$R_{Up}$ encourages the policy to stand up by rewarding the robot’s height $h^{head}_t$. Once the robot has partially risen, $R_{Pitch}$ activates to encourage a vertical torso. This term is gated because it might impede exploration. To discourage thrashing, unsafe motions, or self-collision, small penalties $R_{vel}$, $R_{var}$, $R_{collision}$ are included on abrupt action changes (similar to FRASA~\cite{gaspard2025frasa}). These penalties gently regularize the policy towards smoother, collision-free trajectories without dominating the primary objective.

\subsection{Robust Episode Initialization and Termination}
The initial state of each episode is randomized to simulate arbitrary fallen configurations as seen in FRASA~\cite{gaspard2025frasa}. At the start, the robot is lying with its torso orientation and joint angles displaced up to ± 90° from its initial position. This covers face-up, face-down, and side falls. Each episode runs for a maximum of 10 seconds of simulated time. Episodes are terminated early if the robot enters an unrecoverable state defined by the torso flipping beyond 135° on the pitch axis or an excessively violent motion (angular velocity > 25°/s), which are the same safety cutoffs used in FRASA~\cite{gaspard2025frasa}. An episode is successful if the robot manages to stand up and remain upright for the remainder of the time.

\subsection{Enhanced Domain Randomization}
An extensive domain randomization scheme based on FRASA~\cite{gaspard2025frasa} is applied to ensure that the learned policy is robust to modeling errors and hardware variability. At the start of each training episode, the physical properties are randomized within realistic ranges as seen in Table \ref{tab:domain_randomization}. By randomizing these factors during training, the policy learns to handle a distribution of different dynamics, which is crucial for successful transfer to real hardware and other robot morphologies.

\begin{table}[ht]
\centering
\begin{minipage}[t]{0.48\linewidth}\centering
\caption{Domain Randomization}
\begin{tabular}{ll}
\hline
\textbf{Parameter} & \textbf{Variation} \\
\hline
Mass and Center of Mass            & $\pm$10\% \\
Ground and Actuator Friction       & $\pm$15\% \\
Battery Voltage (Motor Gains)      & $\pm$10\% \\
Sensor Orientation (IMU Offset)    & $\pm$3° \\
\hline
\end{tabular}
\label{tab:domain_randomization}
% \end{table}
\end{minipage}\hfill%
\begin{minipage}[t]{0.48\linewidth}\centering
% \begin{table}[h]
% \centering
\caption{Key Hyperparameters}
\begin{tabular}{ll}
\hline
\textbf{Hyperparameter}        & \textbf{Value} \\
\hline
Network                 & 512-512-256\\
Learning rate                  & $1 \times 10^{-3}$ \\
Batch size                     & 1024 \\
Discount factor $\gamma$       & 0.99 \\
Target update rate             & 0.01 \\
Parallel environments          & 16 \\
\hline
\end{tabular}
\label{tab:hyperparameters}
\end{minipage}
\end{table}

\subsection{DRL Algorithm and Training Enhancements}
The policy was trained with a similar setup as FRASA~\cite{gaspard2025frasa}, using Soft Actor-Critic (SAC)~\cite{haarnoja2018softactorcriticoffpolicymaximum} augmented with the CrossQ algorithm designed for improved sample efficiency~\cite{bhatt2024crossqbatchnormalizationdeep}. The GPU-accelerated Stable Baselines X framework ~\cite{JMLR} was used to speed up training to the point where each policy, after 600k time steps, took only about 1.5 hours on a PC with an AMD 3900X CPU, a NVIDIA RTX 3090 graphics card and 64 GB of DDR4 RAM. Training was performed with 16 parallel MuJoCo environments, each episode randomly initialized with one of seven robot models. Every policy type was trained on 10 random seeds, and success rates were averaged based on 100 episodes per seed for statistical robustness \cite{DBLP:journals/corr/abs-1709-06560}. The 95\% confidence intervals (CI) were calculated with a 10k iteration bootstrap. To cope with the increased task complexity from the diverse dynamics of multiple robots, the neural network capacity was increased to 3 hidden layers, as can be seen in Table \ref{tab:hyperparameters}.

\section{Experiments \& Analysis}
\begin{figure}[t]
  \centering
  \includegraphics[width=12cm]{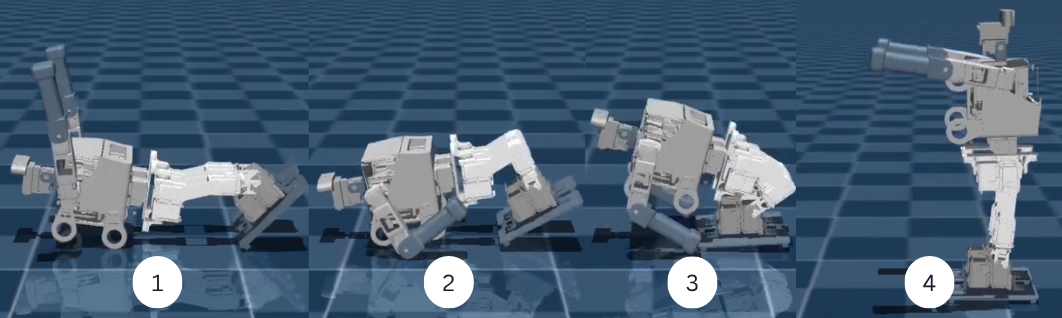}
  \caption{Recovery sequence of the Bez2 robot in Mujoco over 2 seconds.}
  \label{fig:Getup}
\end{figure}

Several experiments were conducted to validate the unified policy and investigate how morphological diversity influences zero-shot generalization. To visualize the learned behavior, Figure~\ref{fig:Getup} shows a full recovery trajectory executed by the shared policy on the Bez2 robot.

% \vspace{-4pt}

\subsection{Leave-One-Out Zero-Shot Generalization}
The leave-one-out (LOO) experiment evaluates whether a shared policy can recover from falls and zero-shot transfer to unseen morphologies without fine-tuning. This indicates that the policy has learned generalizable fall recovery strategies, and not just solutions tailored to the training set. For each of the seven morphologies, 10 random seed shared policies were trained on the other six robots and evaluated on all seven robots. Each policy is tested on 100 episodes per morphology using the standard success criterion: raising the head above its desired height and holding for the remainder of the episode.

\begin{figure}[t]
  \centering
  \includegraphics[width=11cm]{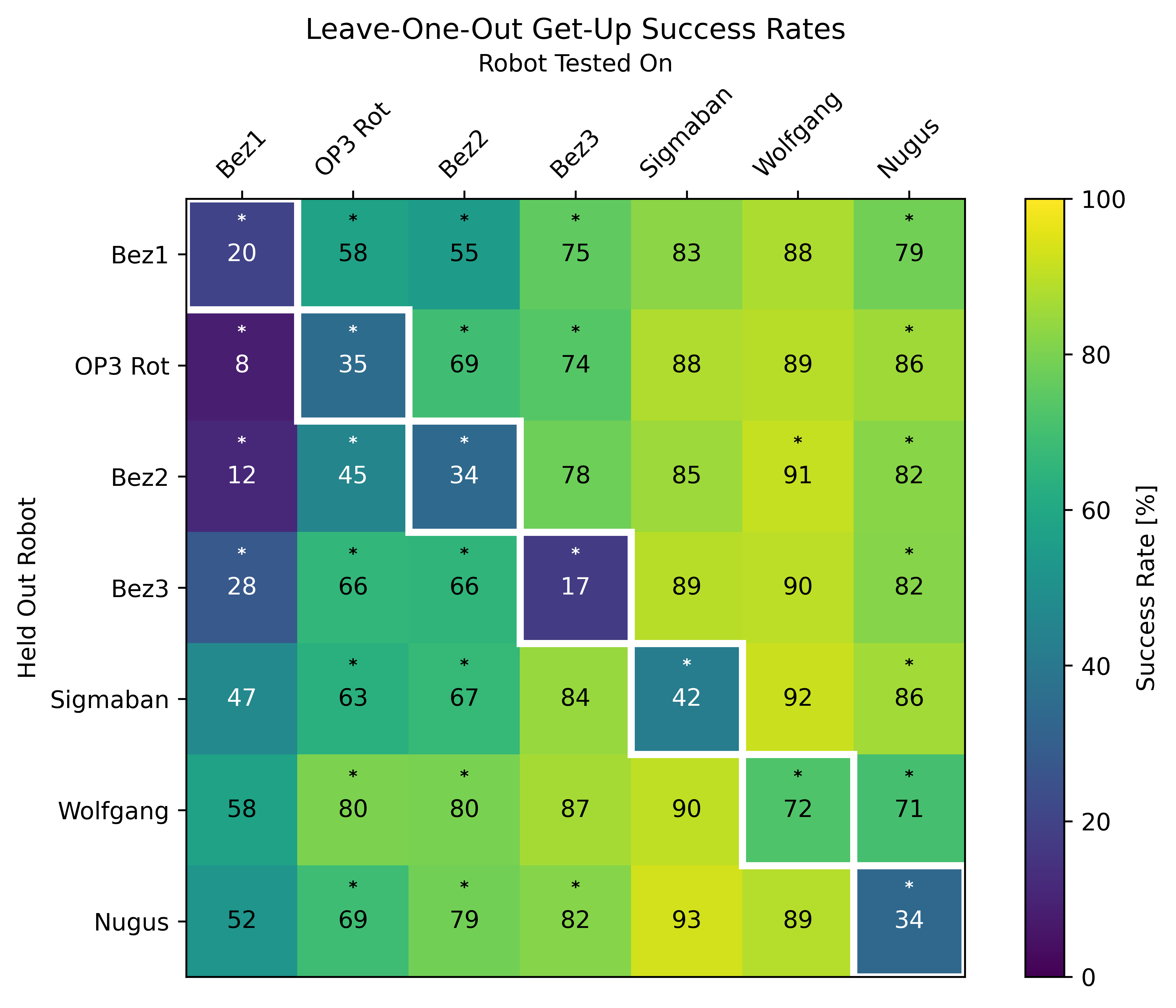}
  \caption{Leave-one-out fall recovery heatmap. Rows: policies trained on six robots (held-out robot excluded); columns: evaluation robot. Cells represent the mean over 10 random seeds (100 episodes each). Thick white boxes mark the zero-shot diagonal; asterisks denote cells with significant differences from the specialized policies (p < 0.05, Welch t-test).}
  \label{fig:loo_heat}
\end{figure}

Figure~\ref{fig:loo_heat} shows an LOO heat-map. Diagonal entries yield zero-shot recovery performance on a robot that was never seen during training, while off-diagonal entries can quantify how sensitive the shared policy is to excluding a specific morphology from training. The shared policy has a great zero-shot transfer to Wolfgang with a mean ± std success rate of 72 ± 21\%; 95\% CI [58, 82], but struggles on the remaining six unseen morphologies, achieving only 17–42\% success rates. These low diagonal values highlight the difficulty of transferring to robots that are top-heavy, long-armed, or otherwise distant in morphology space.

Off-diagonal entries highlight strong robustness, with 21 of 42 entries exceeding 80\%, even after removing one morphology from training. Performance for smaller robots (first two columns) declines when similarly sized robots are omitted. They significantly improve when the tallest robots are excluded. This indicates that certain morphologies offer more transferable experience. In particular, NUGUS could not learn to get up when trained in isolation, yet it benefits strongly from shared training (81 ± 6\%; 95\% CI [77, 84]). This indicates that difficult morphologies benefit from knowledge distilled from shared experience.

This confirms our central hypothesis that one policy can perform fall recovery on most unseen kid-size morphologies, but it also underscores the need to include morphological outliers in the training curriculum to raise the floor on worst-case transfers. For broader generalization (e.g., adult‑size), future work should add curriculum or morphology conditioning.

\subsection{Shared Policy vs. Specialist Policies}
The cost of generalization can be quantified by comparing our best shared policy (trained on 6 robots) to individually trained specialist policies. This experiment also verifies that each morphology is independently learnable and that the observed zero-shot failures in other experiments are not caused by intrinsic difficulty or flawed task setup. For each robot, 10 random seed specialist policies were trained using identical hyperparameters and reward structure as the shared policy. Each policy was evaluated on its corresponding morphology across 100 episodes. Figure~\ref{fig:spec_shared} compares mean success rates (± 95\% CI). Four clear patterns emerge: 

\begin{figure}[t]
  \centering
  \includegraphics[width=12cm]{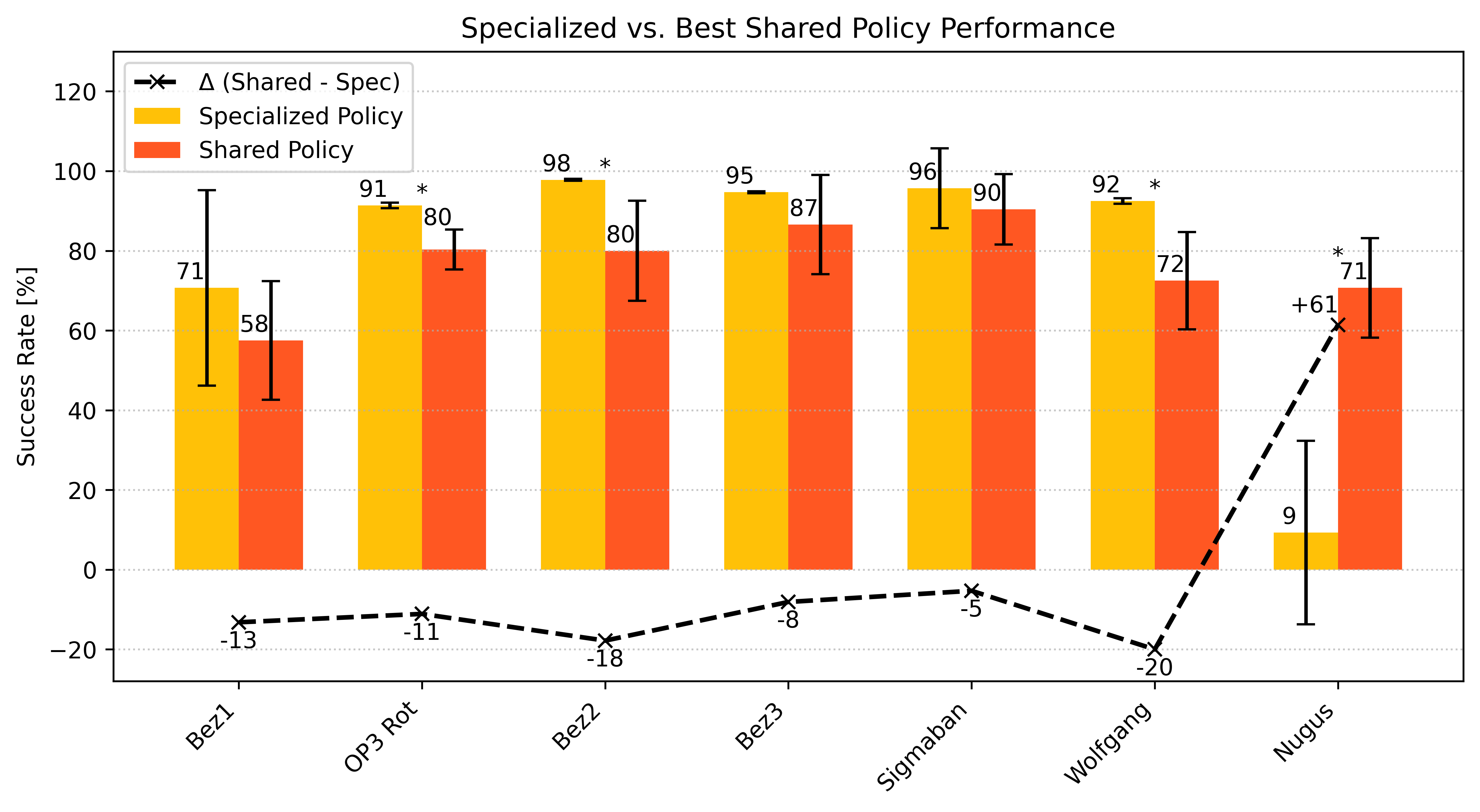}
  \caption{Success rate comparison between the specialists (yellow) and shared (orange) policies. Deltas indicate performance differentials. Error bars denote 95\% CI over 10 random seeds (100 episodes each). Asterisks indicate significant differences from the specialized policies (p < 0.05, Welch t-test).}
  \label{fig:spec_shared}
\end{figure}
% \vspace{-4pt}
% \FloatBarrier 
\begin{itemize}
    \item Large gain on NUGUS: $+\Delta61\%$ (95\% CI [38,85], p < 0.001).
    \item Moderate deficits: Wolfgang $-\Delta20\%$ (p < 0.05) and Bez2 $-\Delta18\%$ (p < 0.05).
    \item Minor deficits: (> $-\Delta15\%$) on Bez1, OP3-Rot, Bez3, and Sigmaban.
    \item Overall robustness: The shared policy still exceeds 58\% success on every robot and surpasses 80\% on four of seven.
\end{itemize}

These results show that sharing experience across morphologies entails only modest performance cost while unlocking new behaviors, most notably on NUGUS, where the specialist policy could not perform fall recovery. To our knowledge, this is the first demonstration of cross-robot skill transfer in fall recovery. The cost of generality is offset by the robustness and transferability of skills across the morphology space. These findings reinforce our thesis: a single shared policy offers a scalable and robust alternative to per-robot controllers, particularly for heterogeneous humanoid teams.

\subsection{Morphological Scaling Analysis}
The size of the training set and diversity of included morphologies have a noticeable effect on the zero-shot performance. This experiment highlights whether more morphologies yield better generalization and whether the type of morphologies matters. 10 random seed policies were trained for each group of $k$ morphologies selected. Each group had morphologies that maximized continuous coverage of size, mass, and limb ratio as seen in Table \ref{tab:morph_sets}. This is repeated with $k = 1 ... 6$ training robots and tracks the performance on two held-out morphologies, Sigmaban (Sig) (midsize, moderate difficulty) and Wolfgang (highest LOO zero-shot transfer rate).

\begin{table}[ht]
\centering
\caption{Training Sets Used in Morphological Scaling Analysis}
\label{tab:morph_sets}
\begin{tabular}{lll}
\toprule
\textbf{Count} & \textbf{Sig: Morphs} & \textbf{Wolfgang: Morphs}\\
\midrule
1 & Bez3 & OP3 \\
2 & Bez3, OP3 & Sig, Bez2  \\
3 & Bez3, OP3, Wolfgang & Sig, Bez2, OP3  \\
3-diverse & Bez3, Bez1, NUGUS & Bez3, Bez1, NUGUS \\
4 & Bez3, OP3, Wolfgang, Bez2 &  Sig, Bez2, OP3, Bez1 \\
5 & Bez3, OP3, Wolfgang, NUGUS, Bez2 & Bez2, OP3, Bez1, NUGUS \\
6 & Adds Bez1 & Adds Bez3 \\
\bottomrule
\end{tabular}
\end{table}

\begin{figure}[t]
  \centering
  \includegraphics[width=10cm]{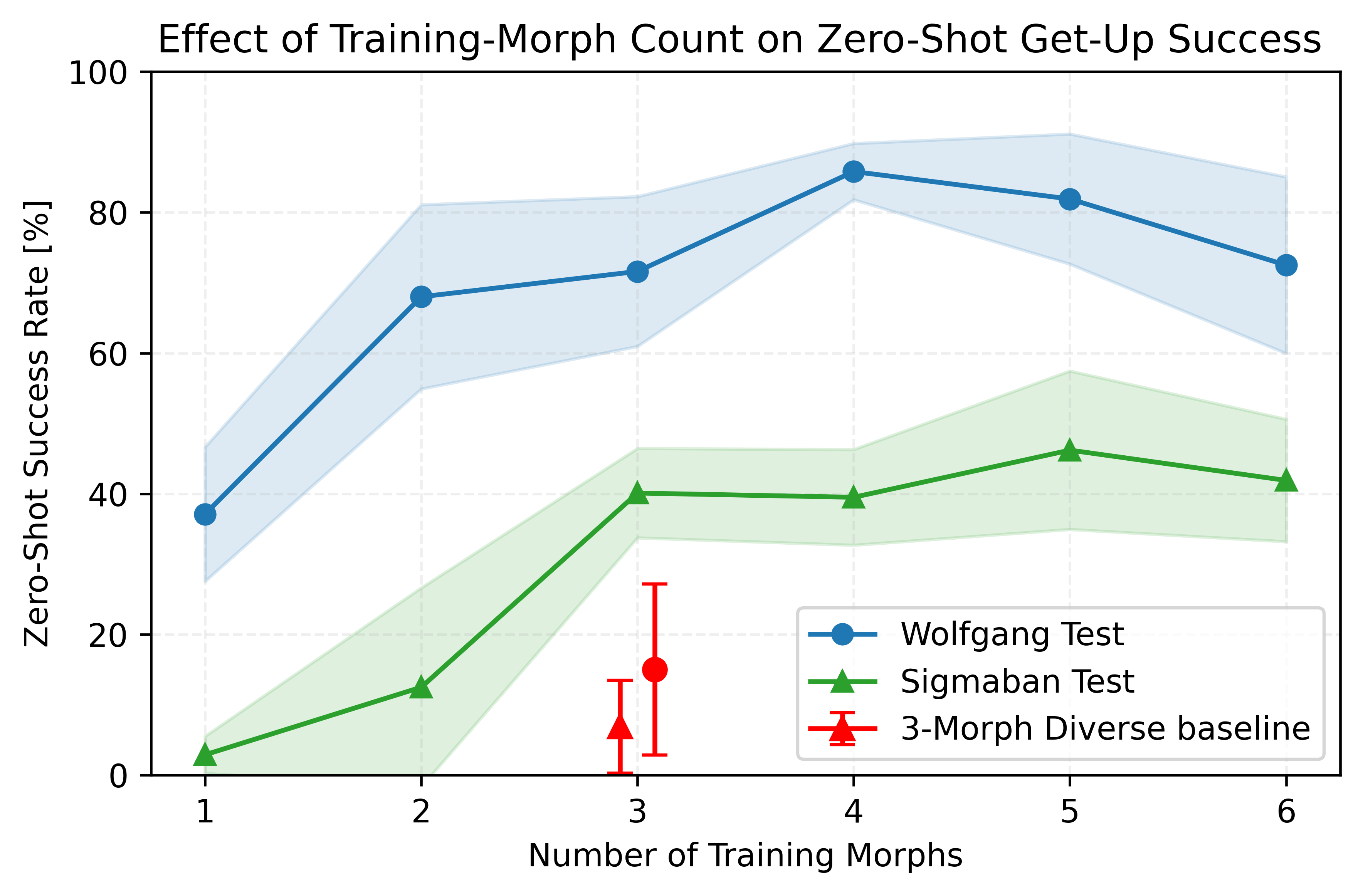}
  \caption{ Zero-shot get-up success rates on Wolfgang (\filledredcirc , blue) and Sigmaban ($\triangle$, green) versus number of training morphs $k$. Shaded bands are 95\% confidence intervals over 10 seeds (100 episodes each). The red marker denotes a purposely diverse three-morph set lacking intermediate morphs. Continuous, overlapping coverage yields steady gains up to $k\approx4-5$; arbitrary diversity alone is insufficient.
  }
  \label{fig:morph_scaling}
\end{figure}
% \vspace{-4pt}
% \FloatBarrier 

Figure~\ref{fig:morph_scaling} plots zero-shot success as a function of training set size and composition. Wolfgang (\filledredcirc, blue): Performance climbs from 37 ± 16\% (95\% CI [28, 47]) with a single training robot to 86 ± 7\% (95\% CI [81, 89]) when four morphs are used, then levels off and dips slightly at k = 6. The drop coincides with adding difficult morphs like NUGUS and Bez3, suggesting that these outliers introduce competing get-up strategies and dilute training time on the morphs that transfer to Wolfgang. Sigmaban ($\triangle$, green): With one or two training robots, success is < 15\%. Introducing more well-chosen morphs raises success to $\approx 40-46\%$, confirming that Sigmaban’s dynamics require broader, overlapping coverage. As soon as difficult or poorly compatible morphs (NUGUS, Bez1) entered the training set, the mean stagnated and the confidence interval widened, reflecting inconsistent learning. A set composed only of extreme morphologies (Bez3, Bez1, NUGUS) achieves $\leq 16\%$ success rate, underscoring that disjoint diversity, without intermediate morphs, fails to generalize.

These results show that generalization depends not just on the number of morphologies, but on which ones are included. Morphological diversity improves zero-shot transfer only when it provides continuous, overlapping coverage of the morphology space. This supports our core thesis: effective generalization requires morphology-aware diversity, not just more robots, but the right ones.

\subsection{Limitations \& Future Baselines}
An ideal baseline would encode morphology data into the policy like NerveNet \cite{wang2018nervenet} or FRASA’s single-robot fall-recovery agent \cite{gaspard2025frasa}. Instead, our specialist vs. shared evaluation serves to benchmark our approach against per-robot policies. Multi-humanoid fall recovery has not yet been shown, and porting existing algorithms to seven morphologies is a non-trivial engineering effort. Therefore, a direct comparison with morphology-aware methods is left to future work. Crucially, our morphology-agnostic training already generalizes across seven kid-size robots, suggesting that implicit dynamics cues can suffice within this morphology range.

\section{Conclusion}
This paper presents a unified reinforcement learning policy that enables zero-shot fall recovery across diverse humanoid morphologies. Unlike prior morphology-agnostic work focused on locomotion (e.g., URMA ~\cite{bohlinger2025policyrunallendtoend}, ModuMorph \cite{xiong2023universalmorphologycontrolcontextual}), our method tackles the more dynamic and complex task of fall recovery \cite{mdpi2023robocup}, achieving robust generalization to unseen robots.

Through leave-one-out zero-shot transfer, policy comparison, and scaling studies, this study showed that a single shared policy can match or even exceed the performance of specialist policies, particularly on morphologies that failed in isolation. These findings highlight that both the quantity and diversity of training morphologies are crucial for generalization.

A key direction for future work is deploying the learned policy on physical humanoid robots. Extensive domain randomization (varying dynamics, frictions, etc.) has been employed to facilitate sim-to-real transfer, and testing of the policy on hardware is in progress.

These results move humanoid control toward morphology-agnostic "generalist" skills, significantly reducing per-robot engineering effort and accelerating the deployment of new robotic platforms.



\bibliography{ref}
\section*{Acknowledgments}
Thank you to the University of Toronto Robotics Association (UTRA) team, especially Alicia Jin. Also, the RoboCup humanoid teams provided models and support.





% LNCS Homepage, \url{http://www.springer.com/lncs}, last accessed 2023/10/25
% \end{thebibliography}
\end{document}